\documentclass[conference]{IEEEtran}

\usepackage{times}

\usepackage[numbers]{natbib}
\usepackage{multicol}
\usepackage[bookmarks=true]{hyperref}

\usepackage{graphicx}
\usepackage{amsmath}
\usepackage{amssymb}
\usepackage{booktabs}
\usepackage{url}
\usepackage{makecell}
\usepackage{floatrow}
\usepackage{epsfig}
\usepackage{graphicx}
\usepackage[titletoc,title]{appendix}
\usepackage{comment}
\usepackage{array}
\usepackage{multirow}
\let\labelindent\relax
\usepackage{enumitem}
\usepackage{wrapfig}
\usepackage{graphicx}
\usepackage{lipsum}
\usepackage{tabularx}
\usepackage{booktabs}
\usepackage{wrapfig,lipsum,booktabs}
\usepackage{caption} 

\usepackage{listings}
\usepackage{algorithm}
\usepackage{algpseudocode}
\usepackage{tabularray}
\usepackage{subcaption}
\usepackage{lipsum}

\newcommand{\ourtitle}{\textbf{Energy-based Models are Zero-Shot Planners for Compositional Scene Rearrangement}}

\newcommand{\model}{\text{SREM}}

\newcommand{\VLMGround}{\text{\texttt{VLMGround}}}

\newcommand{\LLMplanner}{\text{LLMplanner}}

\newcommand{\cliport}{\text{CLIPort}}

\newcommand{\revision}[1]{{\color{black}#1}}

\usepackage[capitalize]{cleveref}
\crefname{section}{Sec.}{Secs.}
\Crefname{section}{Section}{Sections}
\Crefname{table}{Table}{Tables}
\crefname{table}{Tab.}{Tabs.}

\begin{document}

\title{Energy-based Models are Zero-Shot Planners for Compositional Scene Rearrangement}

\author{
\parbox{\linewidth}{\centering
		 Nikolaos Gkanatsios$^{\dagger}$, Ayush Jain$^{\dagger}$,
  Zhou Xian, Yunchu Zhang, Christopher Atkeson, Katerina Fragkiadaki\\
		Carnegie Mellon University
		\\
		{\tt \small{\{ngkanats,ayushj2,xianz1,yunchuz,cga\}@andrew.cmu.edu, katef@cs.cmu.edu}}}
	}

\maketitle
\let\thefootnote\relax\footnotetext{$^{\dagger}$Equal contribution}

 \begin{abstract}
Language is compositional; an instruction can express multiple relation constraints to hold among  objects in a scene that a robot is tasked to rearrange. Our focus in this work is an instructable  scene-rearranging  framework that  generalizes to  longer instructions and to spatial concept compositions never seen at training time. 
We  propose to represent language-instructed spatial concepts with energy functions over relative object arrangements. 
A language parser  maps instructions to  corresponding energy functions and an open-vocabulary visual-language  model  grounds their   arguments  to relevant objects in the  scene. We generate goal scene configurations by gradient descent on the sum of  energy functions, one per language predicate in the instruction.  Local vision-based policies  then re-locate objects to the inferred goal locations. 
We test our model on established instruction-guided manipulation benchmarks, as well as benchmarks of compositional instructions we introduce.  We show our model can execute highly compositional instructions zero-shot in simulation and in the real world.  It outperforms language-to-action  reactive policies and  Large Language Model planners by a large margin,  especially for long  instructions that involve  compositions of multiple spatial concepts. Simulation and  real-world robot execution videos, as well as  our code and datasets are publicly available on our website: 
\url{https://ebmplanner.github.io}.
\end{abstract}
\IEEEpeerreviewmaketitle
  \section{Introduction} \label{sec:intro}

\begin{figure}[t]
  \centering
\includegraphics[width=\linewidth]{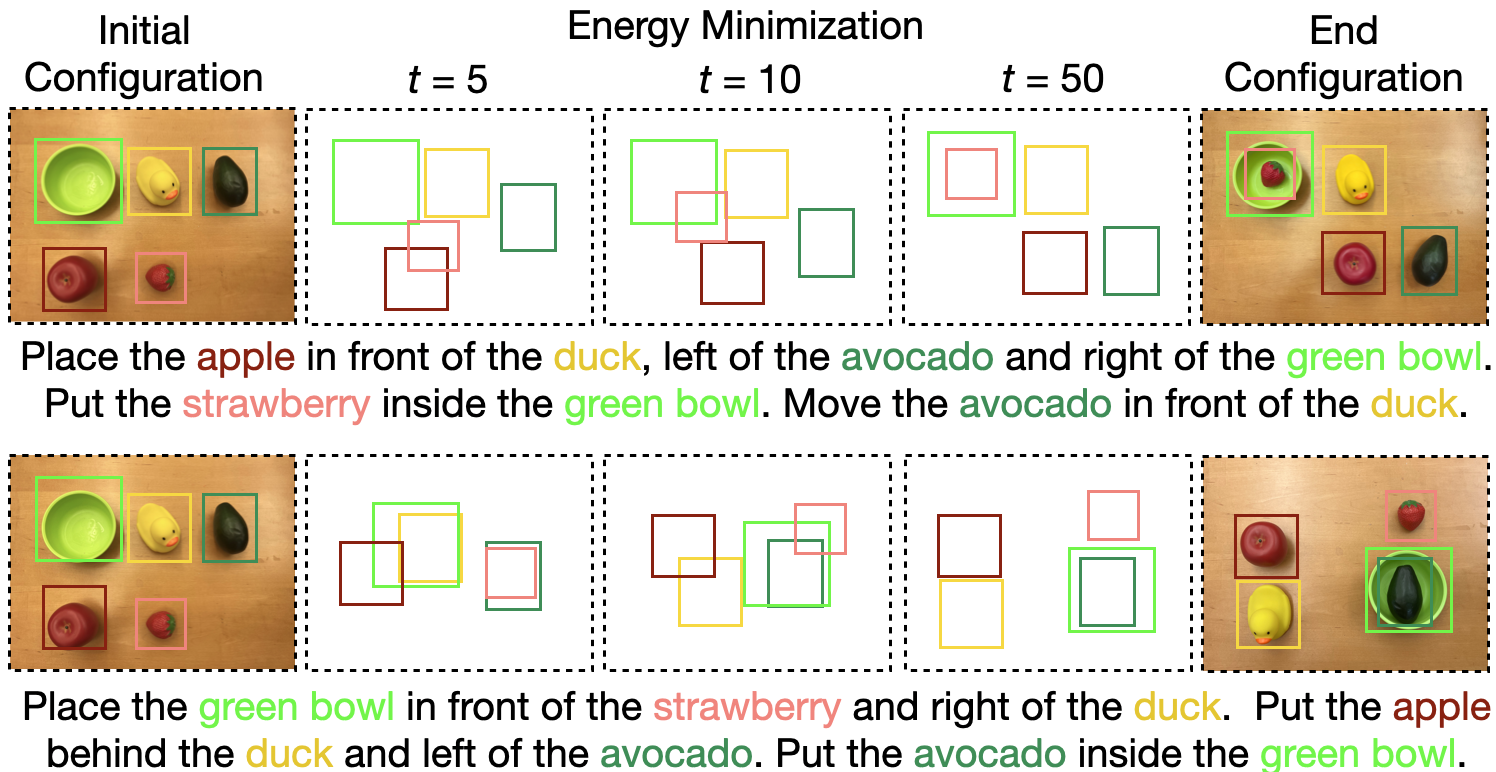}
  \caption{
  \ourtitle.  
  We represent language concepts with energy functions over object locations and sizes. Gradient descent  on the sum of energy functions, one per predicate in the instruction, iteratively updates the object spatial coordinates and generates a goal scene  configuration that satisfies the instruction, if one exists. 
  }
  \label{fig:teaser}
\end{figure}

We consider the scene arrangement task shown in Figure \ref{fig:teaser}. Given a visual scene and an instruction regarding object spatial relations, the robot is tasked to rearrange the objects to their instructed configuration.   Our focus is on strong generalization to longer instructions with novel predicate compositions, as well as  to   scene arrangements that involve  
novel objects and  backgrounds. 

We propose 
generating  goal scene  configurations corresponding to language instructions by  minimizing  a composition of energy functions over object spatial locations, where each energy function corresponds to a language concept (predicate) in the instruction. We  represent each language concept as an n-ary energy function  over relative object poses and other static attributes, such as object size. 
We  train these predicate energy functions  to optimize object poses  starting from randomly sampled object arrangements through Langevin dynamics minimization \cite{EBMimproved}, using a handful of examples of visual scenes paired with  single predicate captions. Energy functions can be 
binary for two-object concepts such as \textit{left of} and \textit{in front of}, or multi-ary for  concepts that describe arrangements for sets of objects, such as \textit{line} or \textit{circle}.  We show that gradient descent on the sum of predicate energy functions, each one involving different subsets of objects, generates a configuration that jointly satisfies all predicates, if this configuration exists, as shown in Figure \ref{fig:teaser}. 

\begin{figure*}[t!]
  \centering
  \includegraphics[width=\linewidth]{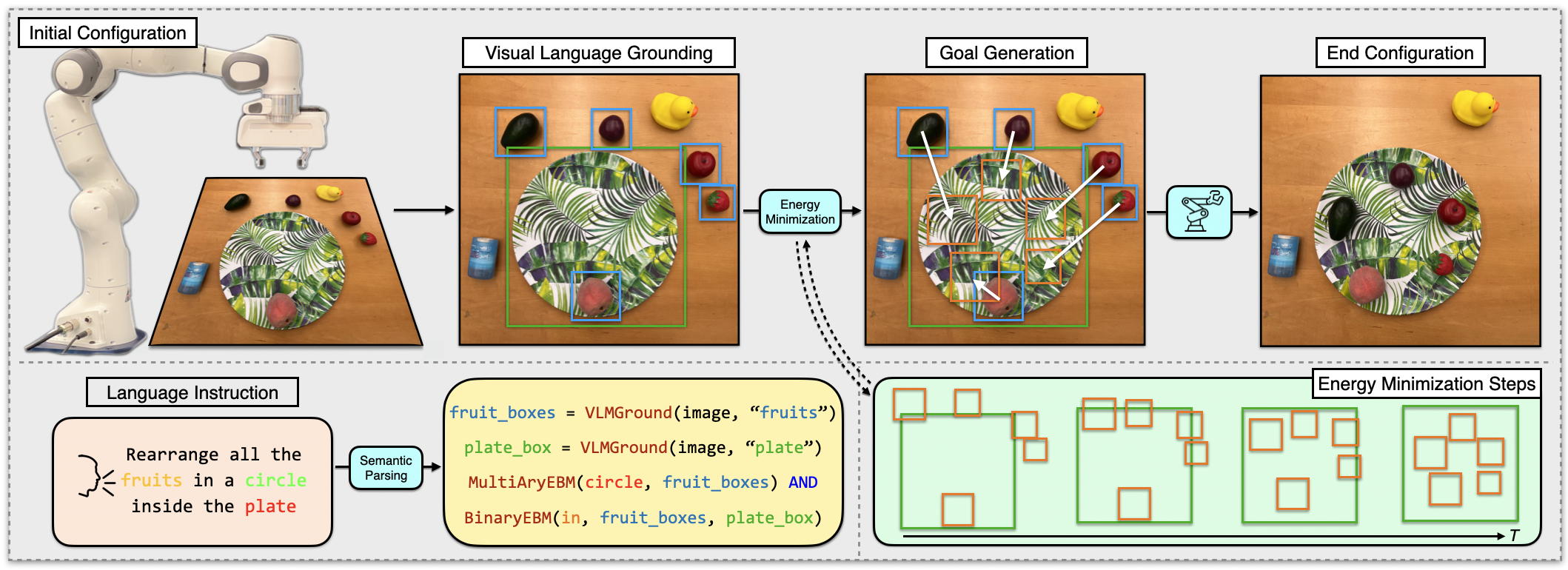}
  \caption{\textbf{Scene rearrangement through energy minimization.} 
  Given an image and a language instruction, a semantic parser maps the language into a set of  energy functions (\texttt{BinaryEBM}, \texttt{MultiAryEBM}), one for each spatial predicate in the instruction, and calls  to an open-vocabulary visual language grounder (\texttt{VLMGround}) to localize the object arguments of each energy function mentioned in the instruction, here ``fruits" and ``plate". Gradient descent on the sum of energy functions with respect to object spatial coordinates generates the goal scene configuration. Vision-based neural policies  condition on the predicted pick and place visual image crops and predict accurate pick and place locations to manipulate the objects. 
  }
  \label{fig:intro}
\end{figure*}

We propose a robot learning framework that harnesses minimization of compositions of energy functions to generate instruction-compatible object  configurations for robot scene rearrangement. A neural semantic parser is trained to map the input  instruction to a set of predicates and corresponding energy functions, and an open-vocabulary visual-language grounding model \cite{DBLP:journals/corr/abs-2104-12763,jain2022bottom}  grounds their arguments to objects in the  scene, as shown in Figure \ref{fig:intro}.  Gradient descent on the sum of energies with respect to the objects' spatial coordinates  computes the final object locations that best satisfy the set of spatial constraints expressed in the instruction. 
Given the predicted  object goal locations, we use vision-based pick-and-place  policies that condition   on the visual patch around the predicted pick and place locations  to rearrange the objects \cite{transporter}. We call our framework Scene Rearrangement via Energy Minimization (\model).

We test \model{} in scene rearrangement of  tabletop environments on simulation benchmarks of previous works \cite{cliport}, as well as on new  benchmarks we contribute that involve compositional instructions. We curate multiple train and test splits to test out-of-distribution generalization with respect to (i) longer instructions with more predicates,  (ii) novel objects and (iii) novel background colors. We show \model{} generalizes zero-shot to complex predicate compositions, such as  \textit{``put all red blocks in a circle in the plate"} 
\textbf{while  trained from single  predicate examples}, such as \textit{``an apple inside the plate"} and \textit{``a circle of blocks"}. We show \model{} generalizes to  real-world scene rearrangement without any fine-tuning, thanks to the object abstractions it operates on. 
We compare our model against  state-of-the-art language-to-action policies   \cite{cliport} as well as Large Language Model planners \cite{https://doi.org/10.48550/arxiv.2207.05608} and show it dramatically outperforms both, especially for long  complicated instructions. We ablate each component of our model  
and evaluate contributions of perception,  semantic parsing, goal generation and low-level policy modules to performance.

In summary, our contributions are:
    \textbf{(i)} A novel energy-based object-centric planning framework for zero-shot compositional language-conditioned  goal scene generation. 
     \textbf{(ii)} A modular system for instruction-guided robot scene rearrangement that uses semantic parsers, vision-language grounding models, energy-based models for scene generation, and  vision-based policies for object manipulation. 
     \textbf{(iii)} A new instruction-guided scene rearrangement benchmark in simulation with compositional language instructions.
    \textbf{(iv)} Comparisons against state-of-the-art language-to-action policies and LLM planners, and extensive ablations. 

Simulation and  real-world robot execution videos, as well as  our code are publicly available on our website: 
\url{https://ebmplanner.github.io}.

 \section{Related Work} \label{sec:related} 
\textbf{Following instructions for  rearranging scenes}: 
Language is a   natural means of communicating  goals and can  easily describe compositions of actions and arrangements \cite{akakzia:hal-03121146,DBLP:journals/corr/abs-2002-09253,https://doi.org/10.48550/arxiv.2206.01134}, providing more versatile goal descriptions
compared to supplying one or more goal images. The latter requires the task to be executed beforehand, which defeats the purpose of instruction \cite{DBLP:journals/corr/abs-1807-04742,DBLP:journals/corr/abs-1903-03698,deformableravens,https://doi.org/10.48550/arxiv.2202.10765}. 
We group methods in the literature in the following broad categories:
\begin{itemize}
    \item \textit{End-to-end language to action policies}  \cite{lynch2020grounding,cliport,structformer,pmlr-v164-stengel-eskin22a} map instructions to actions or to object locations   directly.  We have found that these reactive policies, despite  impressively effective within the training distribution,  typically do not generalize to longer instructions,  new object classes and attributes or novel backgrounds \cite{structformer,cliport}.

\item \textit{Symbolic planners}  such as  PDDL (Planning Domain Definition Language)   planners \cite{DBLP:journals/corr/abs-1911-04679,5980391, 10.5555/2832415.2832517,DBLP:journals/corr/abs-1811-00090} use predefined symbolic rules and known dynamics models, and  infer  discrete task plans given an instruction with lookahead logic search \cite{5980391,DBLP:journals/corr/abs-1802-08705,DBLP:journals/corr/abs-1911-04679,5980391, 10.5555/2832415.2832517,DBLP:journals/corr/abs-1811-00090}. 
Symbolic planners  assume that each state of the world,  scene goal and intermediate subgoal  
can be sufficiently represented in a logical form, using language predicates that describe object spatial relations. 
These methods predominantly rely on manually-specified symbolic transition rules, planning domains and grounding, which  limits their applicability. 

\item \textit{Large language models (LLMs)}  map instructions to language subgoals \cite{DBLP:journals/corr/abs-2012-07277,DBLP:journals/corr/abs-1909-13072,huang2022language,https://doi.org/10.48550/arxiv.2207.05608} or program policies \cite{LMP} with appropriate plan-like prompts. The predicted  subgoals  interface with low-level short-term policies or skill controllers. 
LLMs trained from Internet-scale text have shown impressive zero-shot reasoning capabilities for a variety of downstream language tasks \cite{https://doi.org/10.48550/arxiv.2005.14165} when prompted appropriately, without any weight fine-tuning \cite{DBLP:journals/corr/abs-2201-11903,DBLP:journals/corr/abs-2107-13586}.  
The scene description is usually provided in a symbolic form as a list of objects present, predicted by open-vocabulary detectors \cite{DBLP:journals/corr/abs-2104-12763}. 
Recent works of \cite{LMP,https://doi.org/10.48550/arxiv.2209.00465} have also fed as input overhead pixel coordinates of objects to inform the LLM's predictions. The prompts for these methods need to be engineered per family of tasks. It is yet to be shown how  the composition of spatial concept functions can emerge in this way. 

\end{itemize}

\textbf{Language-conditioned scene generation}: A large body of work has explored scene generation conditioned on text descriptions \cite{DBLP:journals/corr/abs-1804-01622,DBLP:journals/corr/abs-2102-12092,https://doi.org/10.48550/arxiv.2205.11487,https://doi.org/10.48550/arxiv.2206.10789,https://doi.org/10.48550/arxiv.1511.02793}.  
The work of \cite{Kapelyukh2022DALLEBotIW} leverages web-scale pre-trained models \cite{he2017mask,CLIP,Ramesh2022HierarchicalTI} to generate segmentation masks for each object in the generated goal image. Given an input image, their method generates a text prompt using a captioning model and feeds it to a generative model that outputs a goal image, which is then further parsed into segmentation masks. However, the prompt is limited to contain only names of objects and there is no explicit language-guided spatial reasoning. 
In this work, we seek to make scene generation useful as goal imagination for robotic spatial reasoning and instruction following. 
Instead of generating pixel-accurate images, we generate object configurations by abstracting the appearance of object entities. We show this abstraction suffices for a great number of diverse scene rearrangement tasks.

\textbf{Energy-based models}:
Our work builds upon existing work on energy-based models (EBMs) \cite{Xie2016ATO,EBMclass,EBMCL,EBML2C, EBMVG,EBMimproved,Xie2016SynthesizingDP,EBMplan,Xu2019EnergyBasedCI}. Most similar to our work is that of \cite{EBMCL},  which generates and detects spatial concepts  with   EBMs on images with dots, and \cite{EBMVG,EBML2C}, which demonstrate  composability of image-centric EBMs for generating  face images and images from CLEVR dataset \cite{johnson2017clevr}. 
In this work, we demonstrate  zero-shot composability of  EBMs over object poses instead of images, and showcase their applicability on spatial  reasoning and  instruction following for robotic scene rearrangement.

\section{Method}\label{sec:method}

The architecture of \model{} is shown in Figure \ref{fig:intro}. 
The model takes as input an  RGB-D image of the scene  and a language instruction. 
A semantic parser   maps the  instruction to a set of spatial predicate energy functions  and corresponding referential expressions  for their object arguments. An open-vocabulary visual detector  grounds the arguments of each energy function to actual objects in the scene. 
The goal object  locations are predicted via gradient descent on the sum of  energy functions. Lastly, short-term vision-based pick-and-place policies move the objects to their inferred goal locations. 
Below, we describe each component in detail. 

\textbf{A library of energy-based  models for spatial concepts} 
In our work, a spatial predicate   is represented  by an energy-based model (EBM) that takes as input $x$ the set of objects that participate in the spatial predicate and maps them to a scalar energy value $E_{\theta}(x)$.  An EBM 
defines a distribution over configurations $x$ that satisfy its concept through the  Boltzmann distribution $p_{\theta}(x) \propto e^{-E_{\theta}(x)}$. Low-energy configurations imply satisfaction of the language concept and have high probability. 
 An example of the spatial concept can be generated by optimizing for a low-energy
 configuration through gradient descent on (part of) the input $x$.  
We represent each object entity  by its 2D overhead centroid coordinates and box size. During gradient descent, we only update the center coordinates and leave box sizes fixed. We consider both binary spatial concepts (\textit{in}, \textit{left of}, \textit{right of}, \textit{in front of}, \textit{behind}) as well as multi-ary spatial concepts (\textit{circle}, \textit{line}).

Using an EBM, we can sample configurations from $p_{\theta}$, by starting from an initial configuration $x^0$ and refining it using Langevin Dynamics \cite{ld}:
            \begin{equation} \label{eq:ebm_update}
                x^{k+1} = x^{k} -\lambda \nabla_x E_{\theta}(x^{k}) + \epsilon^k z^k, 
            \end{equation}
where $z^k$ is random noise, $\lambda$ is an update rate hyperparameter and $\epsilon^k$ is a time-dependent hyperparameter that monotonically decreases as $k$ increases. The role of $z^k$ and decreasing $\epsilon^k$ is to induce noise in optimization and promote exploration, similar to Simulated Annealing \cite{Kirkpatrick1984OptimizationBS}. After $K$ iterations, we obtain $x^{-}=x^{K}$. During training, we iterate over Equation~\ref{eq:ebm_update} $K=30$ times, using $\lambda=1$ and $\epsilon_k=5e-3$. During inference, we find that iterating for more, e.g. $K=50$ often leads to better solution. In this case we also linearly decay $\epsilon_k$ to 0 for $k>30$.

We learn the parameters $\theta$  of our  EBM using a contrastive divergence loss that
penalizes energies of examples sampled by the model being lower than energies of ground-truth configuration: 
            \begin{equation} \label{eq1}
                \mathcal{L} = \mathbb{E}_{x^{+} \sim p_D}E_{\theta}(x^{+}) - \mathbb{E}_{x^{-} \sim p_{\theta}}E_{\theta}(x^{-}),
            \end{equation} 
where $x^{+}$ a  sample from the  data distribution $p_D$ and $x^{-}$ a  sample drawn from the learned distribution $p_{\theta}$. We additionally use the KL-loss and the L2 regularization proposed in \cite{EBMimproved} for stable training. 
At test time, compositions of concepts can  be created by 
simply summing energies of  individual constituent concept, as shown in Figures \ref{fig:teaser} and \ref{fig:intro}.

We implement two sets of EBMs,  a BinaryEBM and a MultiAryEBM for binary (e.g., \textit{left of}) and multi-ary (e.g., \textit{circle}) language concepts, respectively. The BinaryEBM expects two object arguments, each represented by its bounding box. We convert the object bounding box  to (top-left corner, bottom-right corner) representation. Then we compute the difference between all corners of the two object arguments and concatenate and feed to a multi-layer perceptron (MLP) that outputs a scalar energy value. Note that the energy function only depends on the relative arrangement of the two objects, not their absolute locations. 
The MultiAryEBM is used for order-invariant concepts of multiple entities, such as shapes. The input is a set of objects, each represented as a point (box center). We subtract the centroid of the configuration from each point and then featurize each object using an MLP. We feed this set of object features to a sequence of four attention layers \cite{vaswani2017attention} for contextualization. The refined features are averaged into an 1D vector which is then mapped to a scalar energy using an MLP. We train a separate EBM for each language concept in our vocabulary using corresponding annotated scenes in given demonstrations. Note that annotated scenes suffice to train the energy functions, kinesthetic demonstrations are not necessary, and in practice each EBM can be trained within a few minutes. 
We provide further implementation details and  architecture diagrams for our EBMs in Section \ref{sec:implement} and Figure \ref{fig:conceptEBMs} of the Appendix. We also visualize the energy landscape for various concepts and combinations in \ref{sec:ebm_landscape} and Figure \ref{fig:ebm_landscape} of the Appendix.

\textbf{Semantic parsing} of instructions into spatial concepts and their arguments.  
Our parser maps  language instructions to  instantiations of  energy-based models and their arguments. It is a Sequence-to-Tree model \cite{dong2016language} with a copying mechanism \cite{gu2016incorporating} which allows it to handle a larger vocabulary than the one seen during training. 
The input to the model is a natural language instruction and the output is a tree.  Each tree node is an operation. The three operations supported are i) \texttt{BinaryEBM} which calls a BinaryEBM from our library, ii) \texttt{MultiAryEBM} and iii) \VLMGround{} which calls the visual-language grounding module. Each node has a pointer to the arguments of the operation, language concepts for EBM calls, e.g., \textit{behind}, and noun phrases for grounding model calls, e.g., \textit{``the green cube''}. Nodes in the parsing tree may also have children nodes, which imply nested execution of the corresponding operations. 
The input utterance is  encoded using a pre-trained RoBERTa encoder \cite{liu2019roberta}, giving a sequence of contextualized word embeddings and a global representation of the full utterance. Then, a decoder is iteratively employed to i) decode an operation, ii) condition on this operation to decode or copy the arguments for this operation, iii) add one (or more) children node(s).  For example, the instruction \textit{``a circle of cubes inside the plate"} is mapped to a sum of energy functions where each object of the multi-ary concept \textit{circle} participates in the constraining binary concept \textit{in}:
\begin{align}
\begin{array}{ll}
   E^{total}=&\texttt{MultiAryEBM}(\text{\textit{circle}}, \VLMGround(``cubes"))\\
   & + \sum_i \texttt{BinaryEBM}(\text{\textit{in}}, x_i,\VLMGround(``plate")),\\
   & x_i \in \VLMGround(``cubes"). 
   \end{array}
\end{align}
We train our semantic parser on the instructions of all training demonstrations of all tasks jointly, as well as on synthesized instructions paired with programs, each with 1-7 predicates,  that we generate by sampling from a grammar, similar to previous works  \cite{mao2019neuro,wang-etal-2015-building}. For more details on the domain-specific language of our parser and the arguments for each operation see  Section \ref{sec:implement} and Table \ref{table:DSL} in our Appendix.

We ground noun phrases predicted by our parser with an off-the-shelf language grounding model \cite{jain2022bottom}, which operates as an open-vocabulary detector. The input is the noun phrase, e.g., \textit{``the blue cube"} and the image, while the output is the boxes of all object instances that match the noun phrase. The open-vocabulary detector has been pre-trained for object detection and referential grounding on MS COCO \cite{DBLP:journals/corr/LinMBHPRDZ14}, Flickr30k \cite{Plummer2015Flickr30kEC} and  Visual Genome \cite{Krishna2016VisualGC}. We finetune the publicly available code of \cite{jain2022bottom} on our training data of all tasks jointly.

\textbf{Short-term vision-based manipulation skills}
We use short-term manipulation policies  built upon Transporter Networks \cite{transporter} to move the obejcts to their predicted locations.  
Transporter Networks  take as input one or more  RGB-D images,  reproject them to the overhead birds-eye-view, and predict  two  robot gripper poses: i) a pick pose and ii)  a pick-conditioned placement pose. 
These networks can model any behaviour that can be effectively represented as two consecutive  poses for the robot gripper, such as pushing, sweeping, rearranging ropes, folding, and so on -- for more details please refer to \cite{transporter}. 

We modify Transporter Networks to take as input a small image RGB-D patch, instead of a complete image view.    Specifically, we  consider as input the image patches around the object pick and object goal  locations predicted by our visual grounding and energy-based minimization modules respectively. In this way, the low-level policies know roughly what to pick and where to place it, and only locally optimize over the best pick location, as well as the gripper's relative rotation, within an object of interest, or placement location, at a particular part of the scene, respectively. 
We show in our ablations (Table \ref{tab:ablation}) that using  learning-based pick-and-place policies helps performance, even if the search space is limited thanks to grounding and goal imagination. We train Transporter Networks from scratch on all our pick-and-place demonstration datasets jointly.

\textbf{Termination of execution}: \model{} generates a goal scene by optimizing the relative poses of the objects mentioned in the instruction. We estimate how many objects should be moved by comparing the detected bounding box (by the language grounding model) and the optimized bounding box (by the EBM). For non-compositional tasks that involve binary concepts, we inject the prior that one object is fixed. Then we take as many actions as the number of objects the EBM moved.

\textbf{Closed-loop execution}:  \model{} first generates a goal scene from the input instruction and then executes it. After execution, we re-detect all relevant objects using our VLM-grounder module to check if they are close to their predicted goal locations. If the re-detected object's bounding box and initially predicted goal bounding box intersect over a certain IoU threshold, we consider the goal to be successfully executed. If we fail to reach the goal, we call again our vision based policies using the current scene configuration. Comparing the post-execution object configuration with the initially imagined goal scene allows to track progress and estimate goal completion as we show in the experimental section and in Section \ref{sec:more_exp} and Table \ref{table:failure_detection} of the Appendix.

    \section{Experiments}\label{sec:experiments}
We test \model{} in its ability to follow language instructions for rearrangement of tabletop scenes  in simulation and in the real world. 
We compare our model against LLM planners~\cite{https://doi.org/10.48550/arxiv.2207.05608} and end-to-end language-to-action policies \cite{cliport}. 
Our experiments aim to answer the following questions:

\begin{enumerate}
\item How does \model{} compare to LLM planners in  predicting scene  configurations from  instructions? (Section \ref{sec:gt_results})

\item How does \model{} compare to   state-of-the-art  language-to-action policies for rearranging scenes? How does their relative  performance change with  varying instruction length and varying amount of training data? (Section \ref{sec:main_results})

\item How does \model{} generalize to novel objects, object colors and background colors, compared to an end-to-end language-to-action model? (Section \ref{sec:generalization})

\item How much do different modules of our framework contribute to performance? (Section \ref{sec:ablations})
\end{enumerate}

\noindent \textbf{Benchmarks:} 
Existing language-conditioned manipulation benchmarks are usually dominated by a single spatial concept like ``inside" \cite{cliport}. To better illustrate the compositionality of spatial concepts, we introduce the following set of benchmarks, implemented with PyBullet:
\begin{itemize}
    \item  \textbf{spatial-relations}, containing single pick-and-place instructions with referential expressions in cluttered scenes with distractors, e.g. \textit{``Put the cyan cube above the red cylinder"}. We consider the relations \textit{left of}, \textit{right of}, \textit{in front of}, \textit{behind}.
    
    \item  \textbf{comp-one-step}, containing compositional instructions  with referential expressions in cluttered scenes with distractors that require one object to be re-located to a particular location, e.g. \textit{``put the red bowl to the right of the yellow cube, to the left of the red cylinder, and above blue cylinder"}.

    \item  \textbf{comp-group}, containing compositional instructions  with referential expressions in cluttered scenes with distractors that require multiple objects to be re-located, e.g., 
\textit{``put the grey bowl above the brown cylinder, put the yellow cube to the right of the blue ring, and put the blue ring below the grey bowl''}.
    
    \item  \textbf{shapes}, containing instructions for making multi-entity shapes (circles and lines), e.g. \textit{``rearrange all red cubes in a circle"}. 
\end{itemize}
We further evaluate our model and baselines on four tasks from the CLIPort benchmark \cite{cliport}, namely \textbf{put-block-in-bowls}, \textbf{pack-google objects-seq}, \textbf{pack-google objects-group} and \textbf{assemble-kits-seq}.

For all tasks we train on either 10 or 100 demos and use the same demos to train all our modules, as discussed in Section \ref{sec:method}. We test on 50 episodes per task, where we vary the instruction and the initial configuration of objects. For \textbf{spatial-relations} and \textbf{shapes} each concept corresponds to a task, while the composition benchmarks correspond to one task each.

\noindent \textbf{Baselines:} We compare \model{} to the following baselines:
\begin{itemize}

\item  \cliport{} \cite{cliport}, a model that takes as input an overhead RGB-D image and an instruction and uses pre-trained CLIP language and image encoders to featurize the instruction and RGB image, respectively; then fuses these with depth features to predict pick-and-place actions using the action parametrization of Transporter Networks \cite{transporter}. The model capitalizes  on language-vision associations learnt by the CLIP encoders. We use the publicly available code of \cite{cliport}. We train one CLIPort model on all tasks of each benchmark, e.g., one model for \textbf{spatial-relations}, a different for \textbf{comp-group} etc. 
Note that the original CLIPort implementation assumes access to oracle success/failure information based on which the model can retry the task for a fixed budget of steps or stop the execution if oracle confirms that the task is completed. We evaluate the CLIPort model without this oracle retry but still with oracle information of how many minimum steps it needs to take to complete the task, so we force \cliport{} to take exactly that number of actions.

\item \LLMplanner{}, inspired by \cite{https://doi.org/10.48550/arxiv.2207.05608}, an instruction-following scene-rearrangement model that uses an LLM to predict a sequence of subgoals in language form, e.g. \textit{``pick the red cube and place it to the right of the blue bowl"}. The generated language subgoals  are fed as input to  language-to-action policies, such as  CLIPort. Scene state description is provided as a list of objects in the scene. \LLMplanner{} does not finetune the LLM but instead uses appropriate prompts so that the LLM adapts its behavior in-context and generates similar statements. The prompts include various previous successful interactions between a human user and the model. We design suitable prompts for our introduced benchmarks and use the LLM to decompose a long instruction into simpler ones (see Figure \ref{fig:prompt} in the Appendix for an example). Then, we feed each generated instruction to a CLIPort model, trained as described earlier. Lastly, for tabletop manipulation tasks in simulation, the LLMPlanner of \cite{https://doi.org/10.48550/arxiv.2207.05608} assumes access to an oracle success/failure detector. The difference in our implementation is that we do not assume any success detector. The execution terminates when all language subgoals have been fed to and handled by \cliport{}.
\end{itemize}

Note that \LLMplanner{} boils down to \cliport{} for non-compositional instructions. As such, we compare with \LLMplanner{} only on \textbf{comp-one-step} and \textbf{comp-group}, both in simulation and real world.

\noindent  \textbf{Evaluation Metrics:} 
We use the following two evaluation metrics: 
(i) \textbf{Task Progress (TP)} \cite{transporter} is the percentage of the referred objects placed in their goal location, e.g. 4/5 = 80.0\% for rearranging 4 out of 5 objects specified in the instruction. 
(ii) \textbf{Task Completion (TC)} rewards the model only if the full rearrangement is complete.  
For the introduced benchmarks we have oracle reward functions that evaluate whether the task constraints are satisfied.

\begin{figure}[t!]
  \centering
  \includegraphics[width= \textwidth]{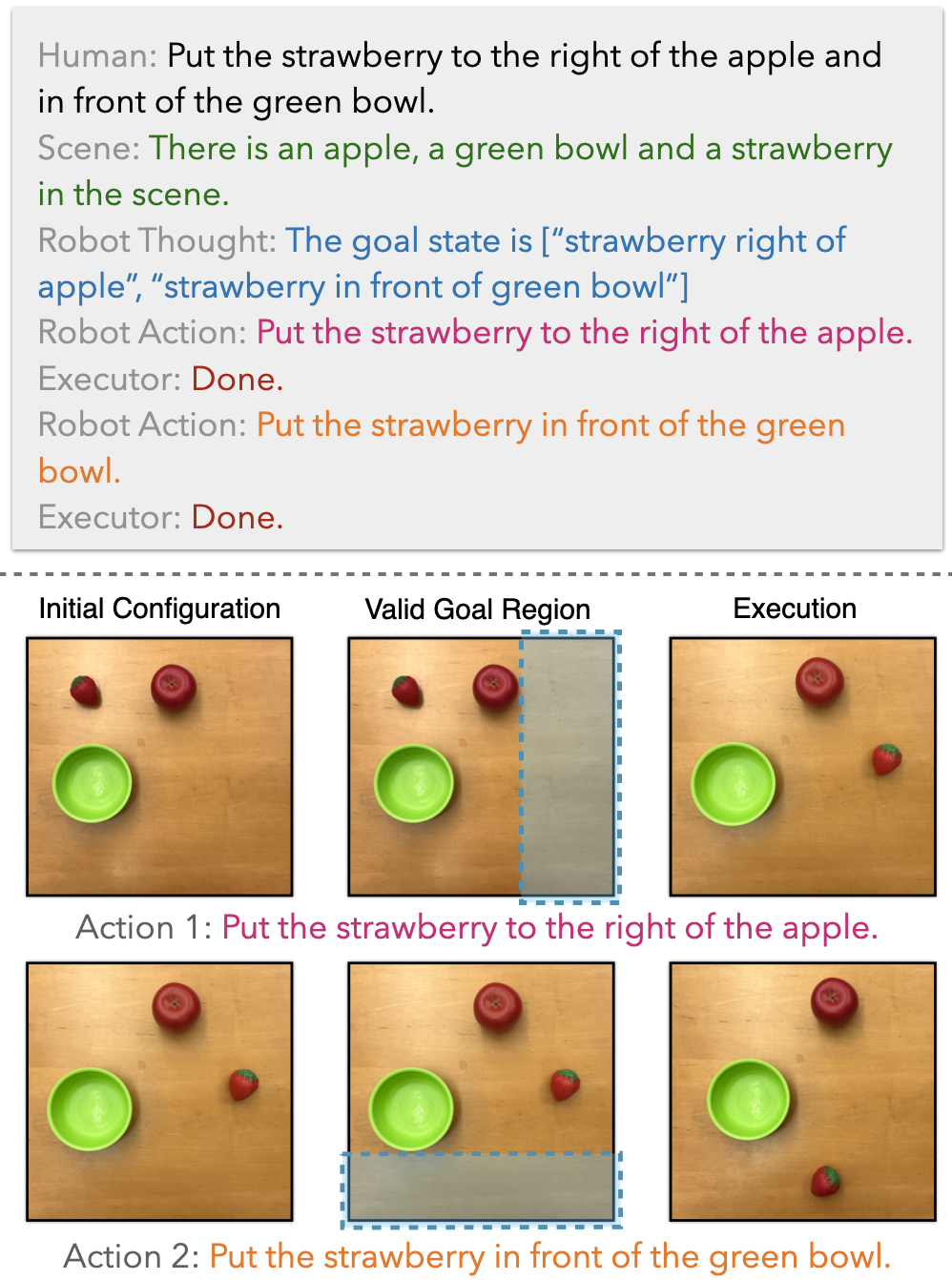}
  \caption{\textbf{Planning in language space with Large Language Models (LLMs).} 
 LLM Planners predict language subgoals that decompose the initial instruction to simpler-to-execute subtasks. Predicted language subgoals are fed to reactive  language-to-action policies for execution.  In cases where concept intersection is needed, the predicted sequential language subgoal decomposition of instructions can  fail. 
 Here, the LLM predicts the first subgoal of  putting the strawberry to the right of the apple.  The reactive policy can succeed if it places the strawberry anywhere within the shaded region. During execution of the next issued language subgoal of putting the strawberry in front of the bowl, the policy violates the first constraint. Placing the strawberry in the intersection of the two shaded regions may not be achieved by decomposing the  two predicates sequentially, as opposed to composing them. Then the burden of handling the compositional instruction is outsourced to the language-to-action policy, which often fails to generalize. Instead, \model{} directly addresses compositionality of multiple spatial language predicates. 
}
  \label{fig:LLMvisual}
\end{figure}
\subsection{Spatial reasoning for scene rearrangement with oracle perception and  control} \label{sec:gt_results}

\begin{table}[t]
\begin{tabular}{lcccccccc} 
\toprule
& \multicolumn{2}{c}{\textbf{comp-one-step}} & \multicolumn{2}{c}{\textbf{comp-group}} \\
Method & TP & TC & TP  & TC \\
\midrule
\LLMplanner{} w/  oracle  & 82.0 & 59.0 & 75.3 & 29.0  \\
\model{}  w/ oracle  & \textbf{90.8} & \textbf{76.0} & \textbf{88.7} & \textbf{62.0} \\
\bottomrule
\caption{\textbf{Evaluation of \model{} and \LLMplanner{} with oracle perception and oracle low-level execution policies} on compositional spatial arrangement tasks. We report Task Progress (TP) and Task Completion (TC).}
\label{tab:oracle}
\end{tabular}
\end{table}

\begin{table*}
\scriptsize
\resizebox{\textwidth}{!}{
\caption{\textbf{Evaluation (TP) of $\model$ and CLIPort on spatial-relations and shapes in simulation.}}
\label{tab:score-spatial-relations}
\begin{tabular}{lcccccccc} 
\toprule
& \multicolumn{2}{c}{\textbf{left-seen-colors}}  & \multicolumn{2}{c}{\textbf{left-unseen-colors}}  & \multicolumn{2}{c}{\textbf{right-seen-colors}} & \multicolumn{2}{c}{\textbf{right-unseen-colors}}  \\
Method & 10 demos  & 100 demos   & 10 demos  & 100 demos  & 10 demos  & 100 demos  & 10 demos   & 100 demos                  \\ 
\midrule
CLIPort & 13.0  & 44.0  & 9.0  & 33.0  & 29.0  & 43.0 & 28.0 & 44.0  \\
\model{} & \textbf{95.0} & \textbf{95.0} & \textbf{93.0} & \textbf{94.0} & \textbf{89.0} & \textbf{92.0} & \textbf{93.0} & \textbf{96.0} \\
\midrule \\
\toprule

& \multicolumn{2}{c}{\textbf{behind-seen-colors}} & \multicolumn{2}{c}{\textbf{behind-unseen-colors}} & \multicolumn{2}{c}{\textbf{front-seen-colors}} & \multicolumn{2}{c}{\textbf{front-unseen-colors}}  \\
Method & 10 & 100  & 10  & 100  & 10 & 100 & 10  & 100  \\ 
\midrule
CLIPort & 24.0  & 45.0  & 22.0  & 51.0   & 23.0  & 55.0   & 13.0 & 40.0  \\
\model{} & \textbf{87.0} & \textbf{87.0} & \textbf{89.0} & \textbf{90.0} & \textbf{89.0} & \textbf{90.0} & \textbf{88.0} & \textbf{89.0} \\
\midrule \\

\toprule
& \multicolumn{2}{c}{\textbf{circle-seen-colors}}  & \multicolumn{2}{c}{\textbf{circle-unseen-colors}}  & \multicolumn{2}{c}{\textbf{line-seen-colors}} & \multicolumn{2}{c}{\textbf{line-unseen-colors}}  \\
Method & 10 demos  & 100 demos   & 10 demos  & 100 demos  & 10 demos  & 100 demos  & 10 demos   & 100 demos                  \\ 
\midrule
CLIPort & 34.1  & 61.5 & 31.2  & 55.6  & 48.6  & 88.2 & 48.6 & 88.5 \\
\model{} & \textbf{91.3} & \textbf{91.5} & \textbf{90.2} & \textbf{ 91.2} & \textbf{98.1} &  \textbf{99.0} & \textbf{98.4} & \textbf{99.4} \\
             
\bottomrule
\end{tabular}}
\end{table*}

\begin{table*}[!t]
\caption{\textbf{Evaluation (TP) of \model{}, \cliport{} and \LLMplanner{} on compositional tasks}. $\model$ is trained only on atomic relations and tested zero-shot on tasks with compositions of spatial relations which involve moving one (\textbf{comp-one-step}) or multiple (\textbf{comp-group}) objects to satisfy all constraints specified by the language. Some language constraints are satisfied already in the initial configuration and the Initial model captures that. 
}
\label{tab:score-compose}
\begin{tabular}{lcccccccc} 
\toprule
& \multicolumn{2}{c}{\textbf{comp-one-step}} & \multicolumn{2}{c}{\textbf{comp-one-step}} & \multicolumn{2}{c}{\textbf{comp-group}} & \multicolumn{2}{c}{\textbf{comp-group}} \\
& \multicolumn{2}{c}{\textbf{seen-colors}} & \multicolumn{2}{c}{\textbf{unseen-colors}} & \multicolumn{2}{c}{\textbf{seen-colors}} & \multicolumn{2}{c}{\textbf{unseen-colors}} \\
\cline{2-9}
Method  & 10   & 100  & 10   & 100  & 10 & 100 & 10  & 100  \\ 
\midrule
Initial (no movement) & 0.0 & 0.0 & 0.0 & 0.0 & 31.7 & 31.7 & 31.8  & 31.8 \\
CLIPort (zero-shot)~ & 9.0  & 12.0 & 7.0  & 12.0  & 37.4 & 37.5 & 32.6 & 38.4 \\
CLIPort  & 13.0 & 15.0 & 14.0 & 9.0  & 38.2  & 38.5 & 34.7 & 40.9\\
\LLMplanner{} & 51.2 & 53.2 & 49.4 & 53.5 & 38.6 & 39.0 & 37.1 & 39.0 \\
\model{} (zero-shot)  & 90.0  & 91.0 & 92.7  & 90.3 & 77.2 & 77.4  & 77.7  & 78.4 \\
\revision{\model{} (zero-shot + closed-loop)}  & \revision{\textbf{91.6}}  & \revision{\textbf{92.0}} & \revision{\textbf{92.9}}  & \revision{\textbf{91.4}} & \revision{\textbf{80.8}}  & \revision{\textbf{81.6}}  & \revision{\textbf{81.1}}  & \revision{\textbf{82.4}} \\
\bottomrule
\end{tabular}
\end{table*}

\begin{table*}[t]
\resizebox{\textwidth}{!}{
\caption{\textbf{Evaluation (TP) of $\model$ and CLIPort on CLIPort benchmark in simulation.}}
\label{tab:cliport-benchmark}
\begin{tabular}{lcccccccc} 
\toprule
& \multicolumn{2}{c}{\textbf{put-block-in-bowl}}  & \multicolumn{2}{c}{\textbf{put-block-in-bowl}}  & \multicolumn{2}{c}{\textbf{packing-google-objects}} & \multicolumn{2}{c}{\textbf{packing-google-objects}}  \\
& \multicolumn{2}{c}{\textbf{seen-colors}} & \multicolumn{2}{c}{\textbf{unseen-colors}} & \multicolumn{2}{c}{\textbf{seq-seen-objects}} & \multicolumn{2}{c}{\textbf{seq-unseen-objects}} \\
Method & 10 demos  & 100 demos   & 10 demos  & 100 demos  & 10 demos  & 100 demos  & 10 demos   & 100 demos                  \\ 
\midrule
CLIPort & 31.0  & 82.1  & 4.8  & 17.6  & 34.8  & 54.7 & 27.2 & 56.4  \\
\model{} & \textbf{84.3} & \textbf{93.8} & \textbf{89.0} & \textbf{95.3} & \textbf{86.8} & \textbf{94.8} &  \textbf{88.0} & \textbf{92.9} \\
\midrule \\
\toprule

& \multicolumn{2}{c}{\textbf{packing-google-objects}} & \multicolumn{2}{c}{\textbf{packing-google-objects}} & \multicolumn{2}{c}{\textbf{assembling-kits}} & \multicolumn{2}{c}{\textbf{assembling-kits}}   \\
& \multicolumn{2}{c}{\textbf{group-seen-objects}} & \multicolumn{2}{c}{\textbf{group-unseen-objects}} & \multicolumn{2}{c}{\textbf{seq-seen-colors}} & \multicolumn{2}{c}{\textbf{seq-unseen-colors}} \\
Method & 10 & 100  & 10  & 100  & 10 & 100 & 10  & 100  \\ 
\midrule
CLIPort & 33.5  & 61.2  & 32.2  & 70.0   & 38.0  & \textbf{62.6}   & 36.8 & \textbf{51.0} \\
\model{} & \textbf{86.1} & \textbf{76.8} & \textbf{87.2} & \textbf{79.6} & \textbf{38.4} & 42.0 &  \textbf{40.8} & 44.0\\
\bottomrule
\end{tabular}}
\end{table*}

In this section, we compare spatial reasoning for predicting compositional scene subgoals in a language space  versus in an  abstract visually grounded space. 
In this section, to isolate this reasoning ability from nuisance factors of visually localizing the objects and picking them up effectively,  we consider \textbf{oracle object detection, referential grounding and low-level pick-and-place policies. } Specifically, we carry out inferred language subgoals from \LLMplanner{} using oracle controllers that relocate an object in the scene such that it satisfies the predicted subgoals.
Note  that \model{} relies on pick-and-place policies that are not language-conditioned, while  \LLMplanner{} relies on language-conditioned  policies for object re-location. Thus, the oracle control assumption is less realistic in the latter case. We forego this difference for the sake of comparison. 

We show quantitative results of \model{} and \LLMplanner{} on the \textbf{comp-one-step} and \textbf{comp-group} benchmarks in Table~\ref{tab:oracle}. Our model outperforms \LLMplanner{} and the performance gap is larger in  more complex instructions. 
To elucidate why an abstract visual space may be preferable for planning, we visualize steps of energy minimization for different instructions in Figure \ref{fig:teaser} and steps of the execution of the LLM prompted by us to the best of our capability in Figure \ref{fig:LLMvisual}. We can see that \model{} trained on single-predicate scenes shows remarkable composability in case of multiple predicates. Language planning on the other hand suffers from the ambiguity of translating geometric concepts to language and vice versa: step-by-step execution of language subgoals does not suffice for the composition of the two subgoals to emerge (Figure \ref{fig:LLMvisual}).

\subsection{Spatial scene rearrangement} \label{sec:main_results}
\textbf{Simulation:} In this section, we compare our model and the baselines in the task of instruction-guided scene rearrangement. We first show results on \textbf{spatial-relations} and \textbf{shapes} in Table \ref{tab:score-spatial-relations}. We largely outperform \cliport{}, especially when less training demos are considered.

To evaluate generalization on longer instructions at test time, we show quantitative results in Table \ref{tab:score-compose} for the benchmarks of \textbf{comp-one-step} and \textbf{comp-group}. We compare our model with \cliport{} trained on atomic spatial relations and zero-shot evaluated on compositional benchmarks. We further fine-tune \cliport{} on demos from the compositional benchmarks. \model{} is not trained on these benchmarks, because the energy functions are already composable, meaning that we can jointly optimize over an arbitrary number of constraints by simply summing the different energy terms. Under all different settings, we significantly outperform all variants of \cliport{} and LLMplanner. We also observe that closed-loop execution boosts our performance further.

We additionally show results on the CLIPort benchmark in Table \ref{tab:cliport-benchmark}. We largely outperform CLIPort on almost all tested tasks. Margins are significantly larger when i) less demos are used and ii) the robot has to interact with objects of unseen colors or classes. Most of the failure cases for our model are due to the language grounding mistakes - in particular for assemble-kits-seq we find that the grounder gets confused between letters and letter holes. 

\textbf{Real World:} We test our model on a  7-DoF Franka Emika robot, equipped with a parallel jaw gripper and a top-down Azure Kinect RGB-D camera. We do not perform any real-world finetuning. Our test set contains  10 language-guided tabletop manipulation  tasks per setting  (Comp-one-step, Comp-group, Circles, Lines).  We show quantitative results in Table \ref{tab:real-world}. \model{} generalizes to the real world without any real-world training or adaptation thanks to the open-vocabulary detector trained on real-world images, as well as the object abstractions in the predicate EBMs and low-level policy modules. We encourage readers to refer to our supplementary video and our website for more detailed results.

\begin{table*}[h!]
\centering
\caption{\revision{\textbf{Generalization experiments of $\model$ and CLIPort in manipulation tasks in simulation (metric is TP).}}}
\label{tab:generalization}
\begin{tabular}{llcccc} 
\toprule
& & \multicolumn{2}{c}{\textbf{spatial-relations}} & \multicolumn{2}{c}{\textbf{composition}} \\ Novel attribute & Model & 10 demos & 100 demos & 10 demos & 100 demos\\
\toprule
\multirow{2}{*}{None} & CLIPort & 22.0 & 47.0 & 25.6 & 26.8 \\ &  \model{}  & 90.0 & 91.0 & 83.6 & 84.2\\
\toprule
\multirow{2}{*}{Color} & CLIPort & 18.0 & 39.0 & 25.1 & 24.5 \\ & \model{}  & 87.0 & 85.0 & 86.5 & 84.0\\
\cmidrule{1-6}
\multirow{2}{*}{Background} & CLIPort & 10.0 & 20.0 & 23.7 & 23.2 \\ & \model{}  & 79.0 & 68.0 & 77.0 & 72.0\\
\cmidrule{1-6}
\multirow{2}{*}{Objects} & CLIPort & 17.0 & 19.0 & 24.5 & 24.8  \\ & \model{}  & 86.0 & 86.0 & 80.9 & 81.5\\
\bottomrule
\end{tabular}
\end{table*}

\begin{table}[t]
\small
    \centering
    \begin{tabular}{lcccc}
        \hline
         Method & \textbf{comp-one-step} & \textbf{comp-group} & \textbf{circles} & \textbf{lines} \\ \hline
         CLIPort & 13.1 & 22.9 & 34.0 & 46.0 \\
         LLMplanner & 39.5 & 25.9 & - & - \\
         \model{} & \textbf{85.6} & \textbf{75.8} & \textbf{94.0} & \textbf{90.0} \\
         \hline
    \end{tabular}
    \caption{\textbf{Real-world evaluation (TP) of \model{}}}
    \label{tab:real-world}
\end{table}

\begin{table}[h!]
\centering
\begin{tabular}{lc}
        \hline
        Method & Accuracy \\ \hline
        \model{} & 77.2 \\
        \model{} w/o goal generation & 42.1 \\
        \model{} w/o learnable policies & 61.2 \\
        \model{} w/ oracle language grounding & 82.3 \\ 
        \model{} w/ everything oracle except goal & 88.3 \\
        \hline
    \end{tabular}
\caption{\textbf{Ablations of  \model{} on the benchmark  comp-group-seen-colors (metric is TP).} \label{tab:ablation}}
\end{table}

\subsection{Generalization analysis} \label{sec:generalization}
We conduct controlled studies of our model's generalization across three axes: a) \textbf{novel colors}: we train the models with objects of 7 different colors and evaluate them on objects of 4 unseen colors; b) \textbf{novel background colors}: we train all models on black-colored tables and evaluate on tables of randomly sampled RGB colors; c) \textbf{novel objects}: we train the models on objects of 4 classes and evaluate on rearrangement of 11 novel classes. In each of these settings, we only change one attribute (i.e. object color, background color or object instance) while keeping everything else constant.

We evaluate our model and CLIPort trained on 10 or 100 demos per task on \textbf{spatial-relations} (average performance over all tasks) and \textbf{composition} (average performance over all tasks from \textbf{comp-one-step} and \textbf{comp-group}). The results are summarized in Table-\ref{tab:generalization}. We observe that our model maintains high performance across all axes of generalization, independently of the number of training demos.

Our model's generalization capabilities rely on the open-vocabulary detector and the fact that EBMs and transporter-based low-level execution policy operate on abstracted space in a modular fashion. While CLIPort models can also generalize to novel scenarios by leveraging the CLIP model, the action prediction and perception are completely entangled and hence even if CLIP manages to identify the right objects based on the language, it has trouble predicting the correct pick and place locations. 

\subsection{Ablations} \label{sec:ablations}
We show an error analysis of our model in Table-\ref{tab:ablation}. First, we remove the goal generation from \model{} (\model{} w/o goal generation) by conditioning the place network on the language input instead of the EBM-generated goal image, while keeping the pick network and object grounders identical. We observe a drop of 35.1\% in accuracy, underscoring the importance of goal generation. We then remove our executor policy (\model{} w/o learnable policies) and instead randomly select pick/place locations inside the bounding box of the relevant object. This results in a drop of 16\%, showing the importance of robust low-level policies. We do not remove the grounder and parser since they are necessary for goal generation. We then experiment with oracle visual language grounder (\model{} w/ oracle language grounding) that perfectly detects the objects mentioned in the sentence, which results in a performance gain of 5.1\%. 
We finally evaluate with perfect grounding, language parsing and low-level execution (\model{} w/ everything oracle except goal) to test the error rate of our goal generator. We obtain an 88.3\% accuracy, thus concluding that our goal generator fails in 11.7\% cases. For a more detailed error analysis, please refer to \ref{sec:more_exp} in the Appendix.

\subsection{Limitations} \label{sec:limits}
Our model presently has the following two  limitations:  First, it predicts the goal  object scene configuration but does not have any knowledge regarding temporal ordering constraints on object manipulation execution implied by physics. For example, our model can predict a stack of multiple objects on top of one another but cannot suggest which object needs to be moved first.  One solution to this problem is to heuristically pick the order based on objects that are closer to the floor in the predicted scene configuration. However, more explicit encoding of physics  priors are important to also identify if the generated configuration is stable or not. A promising direction is to model physics-based constraints as additional energy constraints, and obtain optimization gradients by leveraging either differentiable physics simulators \cite{huang2021plasticinelab, qiao2020scalable, xian2023fluidlab} or learned dynamics models \cite{li2019propagation, xian2021hyperdynamics, pfaff2020learning}. 
Second, our EBMs are currently parametrized by object  locations and sizes, but different tasks need different abstractions. Manipulation of articulated objects, fluids, deformable objects or granular materials,  would require finer-grained parametrization in both space and time. Furthermore, even for rigid objects, many tasks would require finer in-space parametrization, e.g., it would be useful to know a set of points in the perimeter of a plate as opposed to solely representing its bounding box for accurately placing things inside it. Considering EBMs over keypoint or object part graphs \cite{sieb2020graph, manuelli2022kpam} is a direct avenue for future work.

 \section{Conclusion} \label{sec:conclusion}

We introduce \model{}, a modular robot learning framework for instruction-guided scene rearrangement  that maps instructions to object scene configurations via compositional energy minimization over object spatial coordinates.  We test our model in diverse tabletop manipulation tasks in simulation and in the real world. 
Our model outperforms  state-of-the-art end-to-end language-to-action policies, and LLM-based instruction following methods   both in in- and out-of-distribution settings, and across varying amount of supervision. We contribute a new scene rearrangement benchmark that contains more compositional language instructions than previous works, which we make publicly available to the community. Our work shows that a handful of visually-grounded examples suffice to learn energy-based spatial language concepts that can be composed to infer novel instructed scene arrangements, in long and complex compositional instructions.

\section{Acknowledgements} This work is supported by Sony AI, NSF award No 1849287, DARPA Machine Common Sense, an Amazon faculty award, and an NSF CAREER award.

\bibliographystyle{plainnat}
\bibliography{7_refs}

\clearpage
\section{Appendix}

In Section \ref{sec:implement} we give implementation details for the components of our method; in Section \ref{sec:benchmark} we present in more detail the evaluation metrics for the newly-introduced tasks; in Section \ref{sec:more_exp} we show the effectiveness of closed-loop execution for predicting the success or failure of execution and present a detailed error analysis; we show an example prompt used for our LLMPlanner baseline in Section \ref{sec:prompt}; we visualize the learned energy landscapes in Section \ref{sec:ebm_landscape}; in Section \ref{sec:add_rel_w} we include additional related work on constraint-guided layout optimization.

\subsection{Implementation Details}\label{sec:implement}

\begin{figure*}[t]
\begin{center}
    \includegraphics[width=\textwidth]{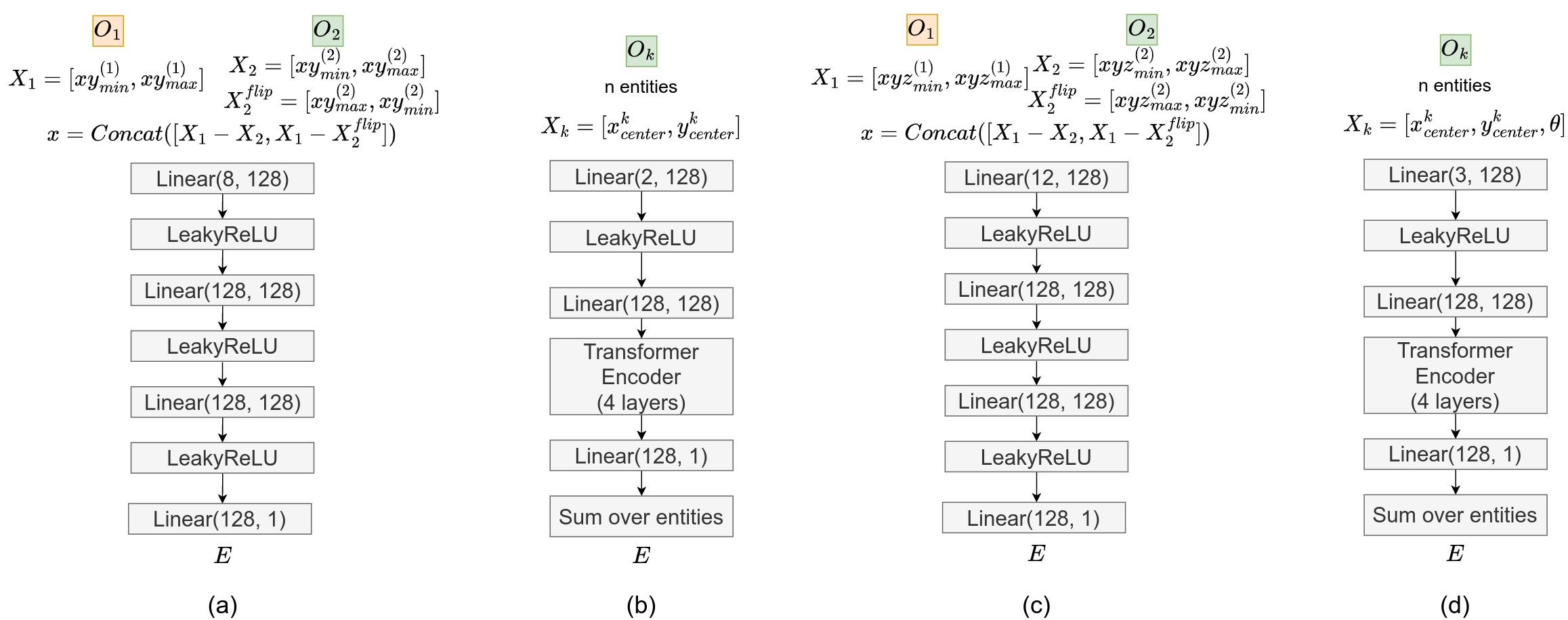}
\end{center}
\caption{\textbf{(a)}: Architecture of the EBM used for binary concepts such as ``right of''. The inputs are two boxes $O_1$ and $O_2$ and the output is the energy of their relative placement. \textbf{(b)}: Architecture of the EBM used for multi-ary concepts such as ``circle''. The input is a set of $n$ entities $O_k, k=1, \dots, n$. The output is the energy of this set of entities wrt the concept. \textbf{(c)}: Architecture of the EBM used for 3D binary concepts such as ``on''. Each object is now represented by a 3D bounding box. \textbf{(d)}: Architecture of the EBM used for concepts that involve pose optimization (rotation). Each object is represented with its center and rotation wrt the global coordinate frame.}
\label{fig:conceptEBMs}
\end{figure*}

\begin{table*}[t]
\caption{All operations in the domain-specific language for \model}
    \begin{tabularx}{0.98\linewidth}{ |l X X|}
        \hline
    \textbf{Operation} & \textbf{Signature} &  \textbf{Semantics}    \\
        \hline
        Filter & (ObjectSet, ObjectConcept) $\rightarrow$ ObjectSet & Filter out set of objects based on some \textit{Object Concept} like object name (eg. cube) or property (color, material) \\
        \hline
        BinaryEBM & (Object A, Object B, Relation) $\rightarrow$ (Pick locations, Place locations) & Executes \textit{BinaryEBMs} for rearranging Object A and Object B to satisfy the given binary \textit{relation} (like left of/right of/inside etc.)  \\
        \hline
        MultiAryEBM & (ObjectSet, Shape Type, Property) $\rightarrow$ (Pick locations, Place locations) & Executes \textit{MultiAryEBMs} for the given \textit{Shape Type} (circle, line, etc.) with specified \textit{Properties} (like size, position etc.) on a set of given objects and generates pick and place locations to complete the shape. \\
        \hline
    \end{tabularx}
    \label{table:DSL}
\end{table*}

\textbf{Energy-based Models}:
The architectures of the BinaryEBM and MultiAryEBM are shown in Figure \ref{fig:conceptEBMs}a and b respectively. We train a separate network for each concept in our library using the same demos that are used to train the other modules. We augment and repeat the samples multiple times to create an artificially larger dataset during training. We use Adam \cite{Kingma2014AdamAM} optimizer, learning rate $1e-4$, batch size 128.


\textit{Buffer}: During training, we fill a buffer of previously generated examples, following \cite{EBMimproved}. The buffer is updated after each training iteration to store at most 100000 generated examples. When the buffer is full, we randomly replace older examples with incoming new ones. We initialize $x_0$ for Equation \ref{eq:ebm_update} by sampling from the buffer 70\% of the times or loading from the data loader 30\% of the times.

\textit{Regularization losses}: We use the KL-loss from \cite{EBMimproved}, $\mathcal{L_{KL}} = \mathbb{E}_{x^{-} \sim p_{\theta}}\bar{E}_{\theta}(x^{-})$, where the bar on top of $\bar{E}$ indicates the stop-gradient operation (we only backpropagate to $E$ through $x^{-}$). We additionally use the L2 energy regularization loss $\mathbb{E}_{x^{+} \sim p_D}E^2_{\theta}(x^{+}) + \mathbb{E}_{x^{-} \sim p_{\theta}}E^2_{\theta}(x^{-})$. We refer the reader to \cite{EBMimproved} for more explanation on these loss terms.

\textit{Extension to tasks with 3D information or pose}:
An important design choice is what parameters of the input we should be able to edit. We inject the prior knowledge that on our manipulation domain the objects move without deformations, so we fix their sizes and update only their positions. Our EBMs operate on boxes so that they can abstract relative placement without any need for object class or shape information. However, EBMs can be easily extended to optimize other types of representations, such as 3D bounding boxes or pose.

We train EBMs that optimize over 3D locations for relative placement. The architecture is shown in Figure \ref{fig:conceptEBMs}c. We adapt the BinaryEBM to represent boxes as $(xyz_{min}, xyz_{max})$ and then compute the relative representations as in the 2D case. We optimize for one 3D relation, ``on". For pose-aware EBMs, we adapt the MultiAryEBM to represent objects as $(x_{center}, y_{center}, \theta)$, where $\theta$ is the rotation wrt the world frame. We then simply change the first linear layer of the MultiAryEBM to map the new input tuple to a 128-d feature vector. The architecture is shown in Figure \ref{fig:conceptEBMs}d. We show qualitative results that compose these EBMs into new concepts on our website.

\textbf{Domain-Specific Language}:
We design a Domain-Specific Language (DSL) which extends the DSL of NS-CL \cite{mao2019neuro} (designed for visual question answering in CLEVR \cite{johnson2017clevr}) to further predict scene generations, e.g. \textit{``put all brown shoes in the green box"}. Detailed description of our DSL can be found in Table~\ref{table:DSL}.

\textbf{Semantic Parser} We construct program annotations for the language instructions of the training demos by mapping them to our DSL (rule-based). We then train our parser on all instructions using Adam optimizer with learning rate $1e-3$ and batch size 32. The same parser weights are used across all tasks. For compositional tasks, we train the parser on the descriptions from the demos that are used to finetune our baselines. The parser is the only part of our framework that needs to be updated to handle longer instructions.

\textbf{Visual-language Grounder}: We finetune BEAUTY-DETR \cite{jain2022bottom} on the scenes of the training demos in simulation. BEAUTY-DETR is an encoder-decoder Detection Transformer that takes as input an image and a referential language expression and maps word spans to image regions (bounding boxes). The original BEAUTY-DETR implementation uses an additional box stream of object proposals generated by an off-the-shelf object detector. We use the variant without this box stream for simplicity. BEAUTY-DETR has been trained on real-world images. We finetune it using the weights and hyperparameters from the publicly available code of \cite{jain2022bottom}.

\textbf{Short-term Manipulation Skills} Our low-level policy network is based on Transporter Networks. Transporter Networks decompose a given task into a sequence of pick-and-place actions. Given an overhead image, the model predicts a pick location and then conditions on it to predict a place location and gripper pose. The original implementation of Transporter Networks supports training with batch size 1 only. We implement a batch-supporting version and find it more stable. We use batch size 8 and follow the original paper in other hyperparameter values.

\subsection{Benchmark Generation} \label{sec:benchmark}
We extend the Ravens \cite{transporter} benchmark for spatial reasoning in the PyBullet simulator. For each benchmark, we write a template sentence (e.g. ``Arrange OBJ1 into a circle") and then randomly select valid objects and colors from a pre-defined list. To test generalization, we include novel colors or novel objects in the evaluation set. Once the sentence is generated, we programatically define valid regions which satisfy the relation and then sample empty locations from it to specify object goal locations. We start by placing all objects randomly in the scene. Then, an oracle hand-designed policy picks and places the objects to the desired locations and returns a demo trajectory which consists of raw RGB-D images and pick-and-place locations. These can be used then to train a behaviour cloning policy similar to CLIPort.

\textit{Evaluation Metrics for Rearrangement Tasks}
To evaluate make-a-circle task, we fit a best-fit circle for the final configuration predicted by \model. To do this, we consider the centers of the bounding boxes as points. Then, we compute the centroid of those points and the distance of each point from the centroid. This is an estimate of ``radius''. We compute the standard deviation of this radius. If this is lower than 0.03, then we assign a perfect reward. The reward linearly decreases when the std increases from 0.03 to 0.06. Beyond that, we give zero reward. We tuned these thresholds empirically by generating and distorting circle configurations.

We follow a similar evaluation strategy for make-a-line. Here we compute the average slope and fit a line to our data. Then we measure the standard deviation of the distance of each point from the line. We found that the same thresholds we use for circles work well for lines as well.

\subsection{Additional Experiments} \label{sec:more_exp}

\textbf{Details on Generalization Experiments}: We conduct controlled studies of our model's generalization across three axes: a) Novel Colors b) Novel background color of the table c) Novel Objects. In each of these settings, we only change one attribute (i.e. object color, background color or object instance) while keeping everything else constant. 
\begin{itemize}
    \item \textit{Novel colors}: We train the models with [``blue", ``red", ``green", ``yellow", ``brown", ``gray", ``cyan"] colors and evaluate them with unseen [``orange", ``purple", ``pink", ``white"]. 
    \item \textit{Novel background colors}: All models are trained with black colored tables and evaluated with randomly sampled RGB color for each instruction.
    \item \textit{Novel objects}: We train the models on [``ring", ``cube", ``cylinder", ``bowl"] and evaluate them with [``triangle", ``square", ``plus", ``diamond", ``pentagon", ``rectangle", ``flower", ``star", ``circle", ``hexagon", ``heart"].
    
    \item \textit{Real-World Experiments} In our real-world experiments, we also use novel objects or novel object descriptions like ``bananas", ``strawberry", ``small objects" which the model hasn't seen during training in simulation.

\end{itemize}

\begin{table}[t]
    \centering
    \caption{\textbf{Performance of failure detection using our generated goal.} We check whether the boxes of the objects in the rearranged scene overlap with the locations the EBM generated. If not, we mark the rearrangement as failed (not satisfying the goal).}
    \begin{tabular}{lccc}
        \hline
        Benchmark & Precision & Recall & Accuracy \\ \hline
        Comp-group & 80.5 & 82.5 & 85.0 \\
        Comp-one-step & 71.4 & 19.2 & 77.0 \\
        \hline
    \end{tabular}
    \label{table:failure_detection}
\end{table}

\begin{figure}[t]
\begin{center}
    \includegraphics[width=\textwidth]{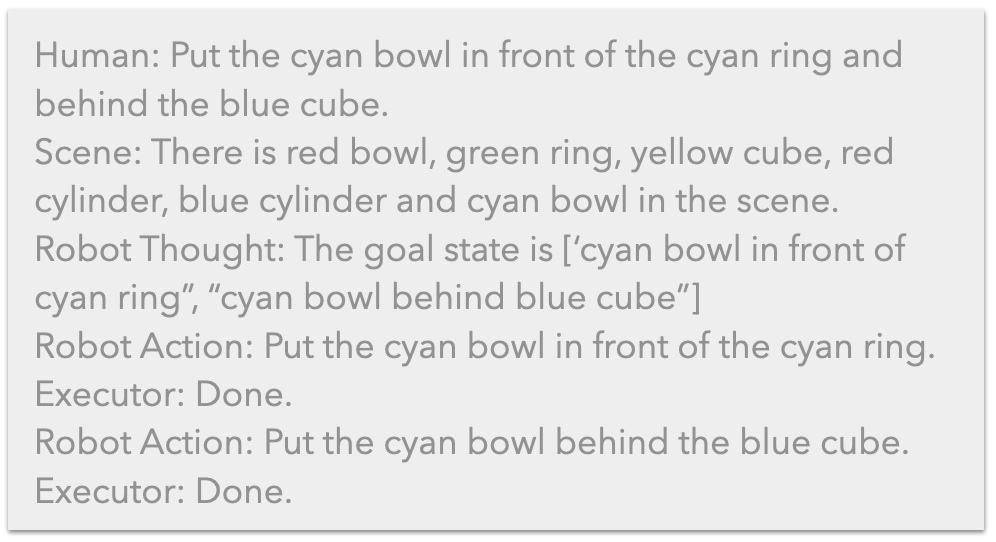}
\end{center}
\caption{\textbf{Example prompt used for LLMPlanner}.}
\label{fig:prompt}
\end{figure}

\textbf{Closed-Loop Execution}:
As we show in the main paper, adding feedback roughly adds 3.5\% performance boost on comp-group benchmark and 1\% boost on comp-one-step benchmark. Retrying cannot always recover from failure, however, it would be still important to know if the execution failed so that we can request help from a human. Towards this, we evaluate the failure detection capabilities of our feedback mechanism in Table-\ref{table:failure_detection}. We observe that all metrics are high for comp-group benchmark. For comp-one-step, however, the recall is very low i.e. only 19.2\%. This is because in this benchmark, we only need to move one object to complete the task and hence most failures are due to wrong goal generation. Thus, the failure classifier classifies those demos as success, because indeed the model managed to achieve its predicted goal and hence results in low recall. This motivates the use of external failure classifiers in conjunction with internal goal-checking classifiers. In contrast, comp-group benchmark is harder because it requires the model to move many objects and thus results in higher chances of robot failures. These failures can be better detected and fixed by our goal-checking classifier. 

Also, note that while previous works like InnerMonologue \cite{huang2022language} show large improvements by adding closed-loop feedback, the improvements for us are smaller. This is because, by design, our model is more likely to reach its goal - since the EBM generates a visual goal and then the low-level policy predicts a pick-and-place location \textbf{within} the predicted goal, it is very likely to satisfy its predicted goal. Indeed, in the comp-one-step, the model satisfies its goal in 93\% cases and in 70\% cases in the comp-group benchmark. In contrast, InnerMonologue does not have any built-in mechanisms for promoting goal-reaching behaviours and thus the difference in performance with additional goal-satisfying constraints is larger for that method.

There is a lot of potential in designing better goal checkers. As already discussed, we can incorporate external success classifiers like those used by prior literature \cite{huang2022language} in conjunction with goal-checking classifiers. Explicit goal generation allows then to check validity of goals directly even before actual robot execution (which makes it safer and less expensive). We leave this for future work. Another promising direction is to add object trackers in the feedback mechanism to keep track of the objects and detect if a failure happened. This is important, if we have multiple objects that are visually similar and hence we would need to keep track of which object corresponds to which goal and retry if it failed to reach it.

\textbf{Error Analysis}:
We conduct a detailed error analysis of our model on comp-group benchmark, shown in Table-\ref{table:error_analysis}. We find that robot failures, i.e. collisions or failure to pick/place objects, result in 6.0\% drop in accuracy. Goal generation adds 11.7\% to the failure. Visual grounding leads to 5.1\% errors while we find language parsing to be nearly perfect.

\begin{table}[t]
    \centering
    \caption{\revision{\textbf{Error Analysis of  \model{} on the benchmark  comp-group-seen-colors.}}}
    \begin{tabular}{lc}
        \hline
        Error Mode & Error Percentage \\ \hline
        Robot failure & 6.0 \\
        Goal Generation failure & 11.7 \\
        Grounding failure & 5.1 \\
        Language Parsing failure & 0.0 \\ 
        \hline
    \end{tabular}
    \label{table:error_analysis}
\end{table}

\begin{figure*}[t]
\centering

   \begin{subfigure}[b]{0.9\textwidth}
   \includegraphics[width=\textwidth]{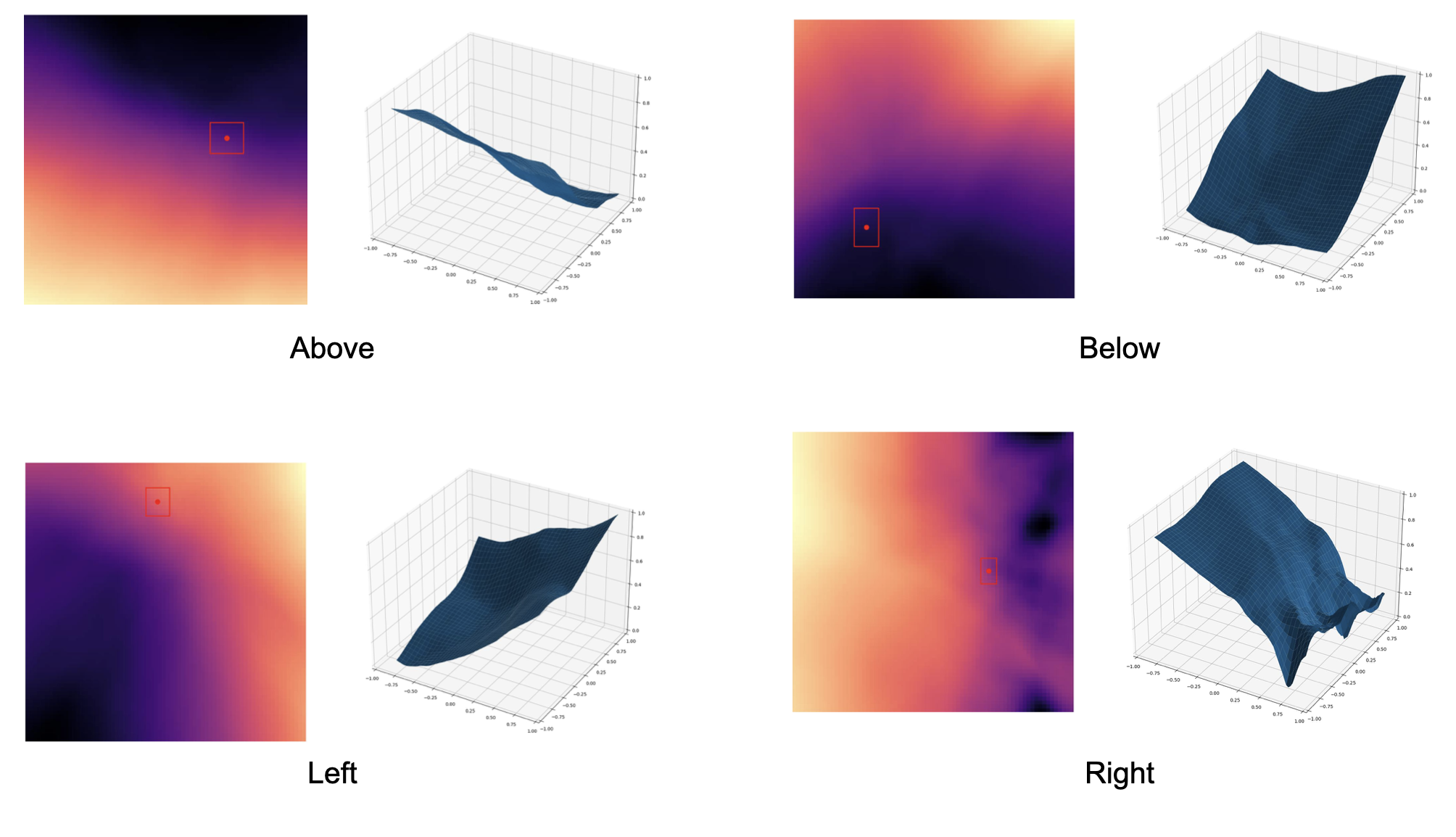}
\end{subfigure}
\begin{subfigure}[b]{0.9\textwidth}
   \includegraphics[width=\textwidth]{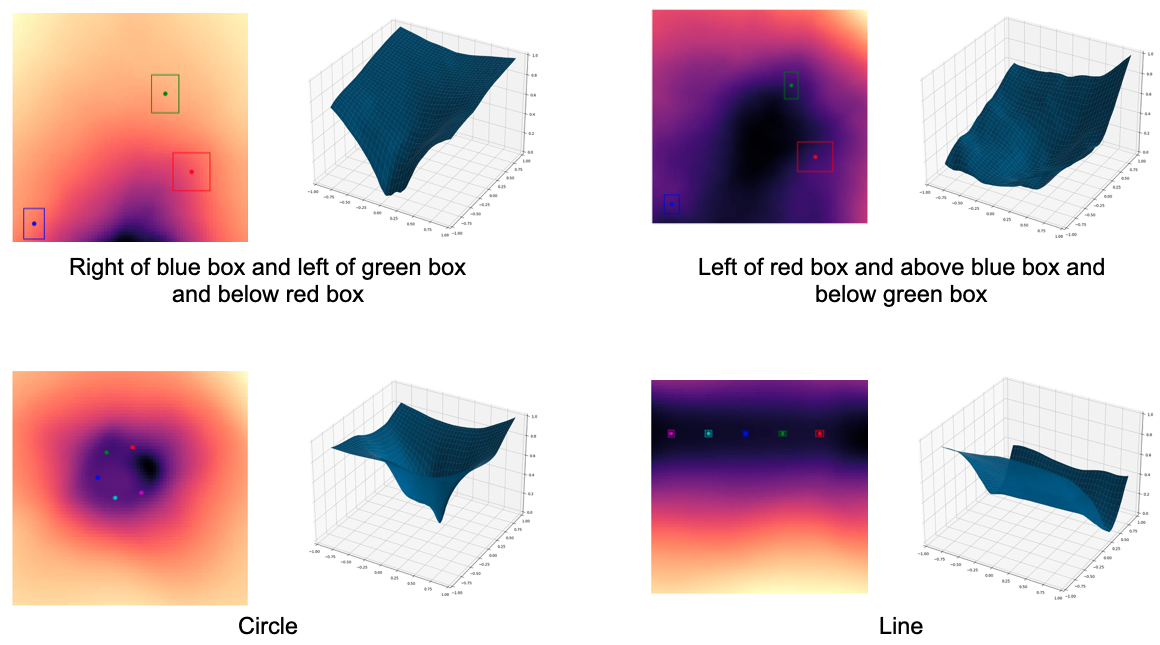}
\end{subfigure}
\caption{\textbf{EBM Energy Landscape visualization:} The boxes shown here remain fixed and we score the energy by moving another box all along the workspace. Energy decreases from white to orange to purple to black.}
\label{fig:ebm_landscape}
\end{figure*}

\subsection{Prompt used for LLMPlanner}\label{sec:prompt}
We show a prompt used for LLMPlanner for \textbf{comp-one-step} in Figure~\ref{fig:prompt}.

\subsection{EBM Energy Landscape Visualization}\label{sec:ebm_landscape}
We visualize the energy landscape of our EBMs for different concepts in Figure-\ref{fig:ebm_landscape}. For binary relations (A, rel, B), we fix B's bounding box in the scene and then move the bounding box of A all over the scene and evaluate the energy of the configuration at each position. For shapes, it is impossible to represent the landscape in 2D or 3D because we need to jointly consider all possible combinations of objects in the scene. Hence, we fix all but one object in a valid circle/line location and move a free box in all possible regions. For compositions of relations, we move only one box and score the sum of energy for all constraints. We expect the energy to be low in regions which satisfy the described concept (relation(s) or shape) and high elsewhere. We observe that the energy landscape is usually smooth with low values in valid regions and high otherwise.

\subsection{Additional Related Work} 
\label{sec:add_rel_w}
\textbf{Constraint-Guided Layout Optimization}: Automatic optimization for object rearrangement has been studied outside the field of robotics. \cite{cons1} and \cite{cons2} use few user-annotated examples of scenes to adapt the hyperparameters of task-specific cost functions, which are then minimized using standard optimization algorithms (hill climbing and/or simulated annealing). To learn those hyperparameters from data, these approaches fit statistical models, e.g. Mixtures of Gaussian, to the given samples. \cite{cons3} further employ such optimization constraints into an interactive environment, where the user can provide an initial layout and the algorithm suggests improvements. All these approaches require expert knowledge to manually design rules and cost function, namely \cite{cons1} identifies seven and \cite{cons3} eleven expert-suggested criteria for successful rearrangement. Since they are hand-crafted, these methods do not generalize beyond the domain of furniture arrangement. In contrast, energy optimization is purely data-driven and domain-agnostic: a neural network scores layouts, assigning high energy to those that do not satisfy the (implicit) constraints and low energy to those who do, essentially modeling the underlying distribution of valid layouts.

\end{document}